\documentclass[sn-mathphys,tikz,border=5pt]{sn-jnl}% Math and Physical Sciences Reference Style
\jyear{2023}%

\usepackage{booktabs} % For formal tables
\usepackage{array}
\usepackage{colortbl}
\usepackage{amsmath,amssymb,amsfonts}
\usepackage{subcaption}
\usepackage{enumitem}
\usepackage{mwe}
\usepackage{graphicx} 
\usepackage{multirow}
\usepackage{flushend}
\usepackage{balance}
\usepackage{soul}
\usepackage{caption} 
\usepackage{verbatim}
\usepackage{tabu}
\usepackage{rotating}

\DeclareMathOperator*{\softmax}{softmax}
\DeclareMathOperator*{\gru}{GRU}
\DeclareMathOperator*{\sigmoid}{sigmoid}

\newcommand{\transpose}{\mathsf{T}}
\newcommand{\matrixize}[1]{\mathbf{#1}}
\newcommand{\eat}[1]{}

\begin{document}

\title[i-Align]{\textit{i-Align}: An Interpretable Knowledge Graph Alignment Model}

\author*[1,3]{\fnm{Bayu Distiawan} \sur{Trisedya}}\email{bayudt@gmail.com}

\author[2,3]{\fnm{Flora D.} \sur{Salim}}\email{flora.salim@unsw.edu.au}

\author[3]{\fnm{Jeffrey} \sur{Chan}}\email{jeffrey.chan@rmit.edu.au}

\author[3]{\fnm{Damiano} \sur{Spina}}\email{damiano.spina@rmit.edu.au}

\author[3]{\fnm{Falk} \sur{Scholer}}\email{falk.scholer@rmit.edu.au}

\author[3]{\fnm{Mark} \sur{Sanderson}}\email{mark.sanderson@rmit.edu.au}

\affil[1]{\orgname{Universitas Indonesia}, \orgaddress{\city{Depok}, \state{West Java}, \country{Indonesia}}}

\affil*[2]{\orgname{University of New South Wales}, \orgaddress{\city{Sydney}, \state{New South Wales}, \country{Australia}}}

\affil[3]{\orgname{RMIT University}, \orgaddress{\city{Melbourne}, \state{Victoria}, \country{Australia}}}

\abstract{Knowledge graphs (KGs) are becoming essential resources for many downstream applications. However, their incompleteness may limit their potential. Thus, continuous curation is needed to mitigate this problem. One of the strategies to address this problem is KG alignment, i.e., forming a more complete KG by merging two or more KGs. This paper proposes i-Align, an interpretable KG alignment model. Unlike the existing KG alignment models, i-Align provides an explanation for each alignment prediction while maintaining high alignment performance. Experts can use the explanation to check the correctness of the alignment prediction. Thus, the high quality of a KG can be maintained during the curation process (e.g., the merging process of two KGs). To this end, a novel Transformer-based Graph Encoder (Trans-GE) is proposed as a key component of i-Align for aggregating information from entities' neighbors (structures). Trans-GE uses \textit{Edge-gated Attention} that combines the adjacency matrix and the self-attention matrix to learn a gating mechanism to control the information aggregation from the neighboring entities. It also uses \textit{historical embeddings}, allowing Trans-GE to be trained over mini-batches, or smaller sub-graphs, to address the scalability issue when encoding a large KG. Another component of i-Align is a Transformer encoder for aggregating entities' attributes. This way, i-Align can generate explanations in the form of a set of the most influential attributes/neighbors based on attention weights. Extensive experiments are conducted to show the power of i-Align. The experiments include several aspects, such as the model's effectiveness for aligning KGs, the quality of the generated explanations, and its practicality for aligning large KGs. The results show the effectiveness of i-Align in these aspects.}


%%\pacs[JEL Classification]{D8, H51}

%%\pacs[MSC Classification]{35A01, 65L10, 65L12, 65L20, 65L70}

\maketitle

\section{Introduction}

Knowledge graphs (KGs) are getting more attention in the research community and industry. They are shown to effectively improve the performance of many downstream applications, such as question answering \cite{berant2013semantic,fader2014open}, recommender systems \cite{zhang2016collaborative,zhao2017meta}, and information retrieval \cite{ensan2017document,liu2015latent,reinanda2020knowledge}. Despite their usefulness, KGs are notoriously incomplete \cite{wang2017knowledge}, and hence they require continuous curation and enrichment.

One of the most effective KG enrichment techniques is KG alignment \cite{paulheim2017knowledge}, which aims to merge two or more KGs to form a single and more complete KG. The first step of KG alignment is finding entities representing the same real-world entity in different KGs. Then, the relationships (i.e., the attributes and the neighbors) of the aligned entities from the different KGs are merged to form a more comprehensive KG. The former step becomes the main challenge of a KG alignment technique, while the latter is straightforward.

Traditional approaches of KG alignment mainly use string matching of entities' attributes to compute entity similarity \cite{volz2009discovering,pershina2015holistic}. These approaches require manually defined constraints, i.e., they need to know which attributes are to be compared beforehand. However, the manually defined constraints are typically sub-optimal since different entities may have different attributes, e.g., a person may have a \texttt{gender} attribute, but an organisation does not.

The state-of-the-art KG alignment approach is based on entity embeddings \cite{trisedya2019entity,wu2020neighborhood,liu2020exploring}. To compute the entity similarity, it first computes the embeddings (a vector representation) of all entities in the KGs. Then, a vector similarity (e.g., cosine similarity) can be used as the entity similarity score. Despite its success, the practical use of embedding-based KG alignment techniques for KG enrichment is low. One of the key reasons is that these techniques do not provide any explanations of the alignment results, which is essential to help experts decide whether the alignment is correct. The expert check is important to maintain the quality of the resulting KG. Without any explanation of the prediction, an expert needs to look up all the attributes and neighbors of the aligned entities predicted by the model to verify the prediction. This can be error-prone since different KGs often have different naming schemes for attributes and relations (e.g., \texttt{educated\_at} vs. \texttt{alumni}).

This paper aims to fill this gap by proposing an interpretable embedding-based KG alignment model capable of generating state-of-the-art performance with explainable alignment results. There are three main challenges to build such a model. \ul{The first is the interpretability of the two common approaches for embedding-based KG alignment techniques, Translation-based and Graph Neural Network (GNN)-based KG alignment models, is non-trivial.} The translation-based alignment models compute entity embeddings using a translation-based embedding model, such as TransE \cite{bordes2013translating}, that treats the triples in a KG independently, making it difficult to compute the importance of attributes and neighbors. Meanwhile, GNN-based models  \cite{wu2020neighborhood,liu2020exploring} overlook entities' attributes by typically only using the entity label to initialize node embeddings in the GNN while ignoring the other attributes (e.g., birth date, address, etc.). Moreover, GNN typically employs a message-passing paradigm, where the aggregation function is constructed to be invariant to neighborhood permutations \cite{dwivedi2021generalization}. These limitations mean that the predicted alignments are difficult to explain, i.e., it is difficult to compute the importance of the attributes and neighbors of the aligned entities.

\ul{The second challenge is that applying a post-hoc (model-agnostic) explainer is sub-optimal}. One of the state-of-the-art post-hoc explainers for GNN models is GNNExplainer \cite{ying2019gnnexplainer}. However, it can only extract the most influential neighbors, but not the most influential attributes, due to the first limitation of GNN-based models mentioned above. Moreover, model-agnostic explainers cannot have perfect fidelity with respect to the model \cite{rudin2019stop}. \ul{The third challenge is scalability}. The state-of-the-art alignment models are built on top of GNN models that need to maintain the whole KB graph, which requires large amounts of memory for the message-passing procedure. This is problematic when handling large KGs.

This paper proposes an interpretable KG alignment model named \textit{i-Align} to handle the above challenges. The main goal of i-Align is to accurately predict entity alignment between KGs and seamlessly provide an explanation for the prediction. The provided explanation is in the form of the similarity between the top-n features (i.e., attributes and neighbors) of the aligned entities used to compute the entity embeddings. Intuitively, a KG alignment model should capture the aligned features of aligned entities, which are reflected by the computed embeddings. In other words, entity embedding is computed by highlighting the features that are aligned with the features of the counterpart entity. Hence, the top-n highlighted attributes and neighbors of entities can help to indicate the correctness of the predictions.

The proposed model is built on top of a Transformer model \cite{vaswani2017attention} to exploit its self-attention mechanism to rank the importance of the attributes and neighbors \cite{wiegreffe2019attention} as the model's explanation. It has two Transformer encoders, one is used as an attribute aggregator, and the other is used as a neighbor aggregator. The attribute aggregator computes a hidden state of an entity based on its attribute using the standard Transformer \cite{vaswani2017attention}. At the same time, the neighbor aggregator computes a hidden state based on its structure/neighbors. Both hidden states are combined to form entity embeddings.

This, however, poses an additional challenge. Computing the hidden states (latent information) based on structure/neighbors using a Transformer-based model is challenging, especially in large KGs, which leads to the following sub-problems. First, the self-attention mechanism of a Transformer is computationally expensive and may not be feasible to be applied to a large graph. Thus, the model needs to decompose the large graph into sub-graphs (mini-batches). Second, it requires a message-passing-like mechanism to aggregate the structural information of both the sub-graphs and the whole graph accordingly. Existing work, such as GraphTransformer \cite{dwivedi2021generalization}, has attempted to simulate the message-passing mechanism of a GNN in a Transformer-based model, but it cannot handle large graphs.

To address the above challenges, a novel Transformer-based Graph Encoder, \textit{Trans-GE}, is proposed for the neighbor aggregator component of i-Align. Trans-GE uses \textit{Edge-gated Attention} that combines the adjacency matrix and the self-attention matrix to learn a gating mechanism to control the information aggregation from neighboring entities. It also uses Historical Embeddings \cite{chen2018stochastic,Fey2021gnnautoscale}, which allows Trans-GE to approximate the full computational graph in a mini-batch to address the scalability issue when encoding a large KG. The attention mechanism of the attributes and neighbor aggregators is used to compute the attention weight to highlight the important attributes and neighbors, respectively. The top-n highlighted attributes and neighbors of the aligned entities are then listed as an explanation of whether the alignment is correct. %Manually, an expert can look into the list to see the similarity of the highlighted attributes and neighbors. Automatically, a set similarity can be used to score the list.

In summary, the contributions of the paper are as follows:

\begin{enumerate} 
	\item An interpretable KG alignment model is proposed, where an explanation of the alignment prediction can be automatically derived. The alignment prediction can help enrich a KG, and the explanation can help experts check the correctness of the prediction to maintain the high quality results of the enrichment process.
	\item Along with the proposed model, a novel Transformer-based graph encoder is proposed. It uses Edge-gated Attention to learn a weight that controls information aggregation from the surrounding neighbors of an entity. It also uses Historical Embeddings to train the model over small mini-batches.
	\item Extensive experiments and analyses are conducted to show the model's effectiveness in predicting the alignments and providing explanations.
\end{enumerate}

\section{Related Work}
\label{sec:related_work}
In this section, first, two approaches of embedding-based KG alignment models are discussed, including translation-based and GNN-based models. Then, explanation techniques for GNN models are also discussed.

\subsection{Embedding-based KG Alignment}
\label{sec:related_work_embedding_based_kga}

Embedding-based KG alignment models have become popular since the success of knowledge representation learning \cite{wang2017knowledge}. Two common approaches of embedding-based KG alignment are translation-based and GNN-based models. Generally, the embedding-based KG alignment models consist of two modules. The first is the embedding module that computes the embeddings of entities and predicates of two KGs. The second is the alignment module that learns an alignment matrix to transform entity embeddings from one KG vector space to the other KG vector space so that the entity embeddings from the two KGs fall in the same vector space. The alignment module is trained using a seed set of aligned entities (\textit{seed alignment} for short). The fundamental difference between the two is how the entity embedding is computed. Translation-based models use a translation-based KG embedding model, such as TransE \cite{bordes2013translating}. In contrast, GNN-based models use a Graph Neural Network encoder, such as Graph Convolutional Networks (GCN) \cite{kipf2017semi}.

\textbf{Translation-based models.} MtransE \cite{chen2017mtranse} is the first effort on embedding-based KG alignment. It uses TransE in its embedding module and shallow networks to learn an alignment matrix. IPTransE \cite{zhu2017iterative} uses an extension of TransE called PTransE, which advances TransE by considering a path between two entities in a KG triple. Another representative KG alignment is BootEA \cite{sun2018bootstrapping}, which iteratively adds the seed alignment from the highly-confident predicted alignments in the previous training iteration. The alignment module of BootEA is a one-to-one classifier instead of a transformation matrix used by the predecessors. TransEdge \cite{sun2019transedge} aims to improve the embedding module by considering the contextual information when computing the predicate embeddings. When processing a KG triple, it uses a function to combine the predicate embeddings with the head and tail embeddings.

The above models only consider the KG structures (entity's neighbors) and ignore the attributes. The subsequent development of embedding-based KG alignment models shows that exploiting the attributes improves the model's performance significantly. JAPE \cite{sun2017cross} is the first to study the integration of attribute information for KG alignment. It treats the attribute as an \textit{attribute triple}, which contains an entity, a relationship/predicate, and an attribute, similar to the typical KG triple. Hence, the TransE embedding models can be directly applied in its embedding module. Further development advances JAPE by using a more robust attribute encoder. For example, AttrE \cite{trisedya2019entity} uses LSTM and N-gram embeddings to encode the attribute values, while MultiKE \cite{zhang2019MultiKE} uses Convolution Neural Networks \cite{lecun2015deep}. The other work uses additional textual information to improve the embedding module. For example, KDCoE \cite{chen2018kdcoe} uses entity description.

% \textbf{Translation-based models.} Earlier KG alignment models, such as MTransE \cite{chen2017mtranse} and IPTransE \cite{zhu2017iterative}, use TransE and its extensions in their embedding module and use shallow networks to learn an alignment matrix. The next developments proposed using additional information, such as unlabelled data via bootstrapping \cite{sun2018bootstrapping} and contextual predicate information \cite{sun2019transedge}. Recent approaches have sought to combine both structural and attribute information \cite{sun2017cross,trisedya2019entity,zhang2019MultiKE,chen2018kdcoe}. Despite their good performance, translation-based models lack transparency, i.e., it is difficult to explain the alignment prediction. This is because they compute entity embeddings using a translation-based embedding model, such as TransE \cite{bordes2013translating}, and its successors \cite{wang2014knowledge,lin2015learning}. TransE computes entity embeddings based on all triples in a KG. A triple in a KG consists of three elements $\langle h,r,t \rangle$, representing a relationship/predicate $r$ between the head entity $h$ and the tail entity $t$. Here, TransE assumes that the embeddings of the head entity can be translated into the embeddings of the tail entity based on the embeddings of the relationship, i.e., $\matrixize{h} + \matrixize{r} \approx \matrixize{t}$. It also that treats the KG triples independently. Thus, the latent connections between entities in the resulting embeddings can not be quantified.

\textbf{GNN-based models.} GCN-Align \cite{wang2018cross} is the first method that uses Graph Convolutional Networks (GCN) as an embedding module. The reason is that GCN can better capture the structural information in a graph (e.g., a KG). HGCN \cite{wu2019jointly} improves GCN-Align by explicitly adding predicate embeddings as additional features. MuGNN \cite{cao2019multi} further improves GCN-Align by inferring the missing relationships/predicates using the Horn rules. GMNN \cite{xu2019cross} treats KG alignment as graph matching and uses node-level and graph-level matching modules. The node-level matching is similar to the embedding module of the predecessors, while the graph-level matching uses another GCN to encode \textit{topic graphs} contained in a KG. NMN \cite{wu2020neighborhood} uses neighborhood differences as additional features. They propose a sampling strategy to select the most representative/informative neighbors to get the neighborhood difference.

The above-mentioned models use GCN in their embedding module. Some models use the variants of GCN, such as KECG  \cite{li2019semi} uses Graph Attention Networks (GAT) \cite{velivckovic2018graph}. AVR-GCN \cite{ye2019vectorized} proposes vectorized relational GCN. SSP \cite{nie2021global} combines TransE and GCN as its embedding module. The other work along this line uses various information as additional features. For example, AliNET \cite{sun2020knowledge} considers the neighbor distance, MRAEA \cite{mao2020mraea} includes meta-relation, such as relation direction and inverse relation, and AttrGNN \cite{liu2020exploring} splits the attributes into different views, including attribute label, attribute literal values, and attribute digital values.

% \textbf{GNN-based models.} GCN-Align \cite{wang2018cross} was the first method to use Graph Convolutional Networks (GCN) as an embedding module. The next development aimed to improve GCN-Align by different strategies, such as adding predicate embedding \cite{wu2019jointly}, inferring the missing relationships/predicates \cite{cao2019multi}, adding graph-level matching \cite{xu2019cross}, and using neighborhood difference \cite{wu2020neighborhood}. Some models use the variants of GCN, such as KECG  \cite{li2019semi} that uses Graph Attention Networks (GAT) \cite{velivckovic2018graph}, AVR-GCN \cite{ye2019vectorized} that proposes vectorized relational GCN, and SSP \cite{nie2021global} that combines TransE and GCN. Other work has explored the use of various information as additional features, for example, AliNET \cite{sun2020knowledge} considers the neighbor distance, MRAEA \cite{mao2020mraea} includes meta-relation information, and AttrGNN \cite{liu2020exploring} splits the attributes into different views.

In general, the explanation of the predicted alignment from GNN-based models is difficult to obtain. The reason is that the GNN used in these models employs a message-passing paradigm, where the aggregation function is constructed to be invariant to neighborhood permutations \cite{dwivedi2021generalization}. Thus, it is difficult to quantify the importance of the neighbors for computing the entity embeddings. Moreover, GNN-based models overlook entities' attributes. Typically, they only use the entity label to initialize node embeddings in the GNN and ignore the other attributes. This makes it difficult to measure the importance of the attributes. 
One alternative to getting an explanation of GNN-based models is by applying a post-hoc explainer, which is discussed in the following section.

\subsection{Explanation Techniques for GNN}
\label{sec:related_work_explanation_gnn}
A post-hoc explainer treats a machine learning model (e.g., a GNN model) as a black box. It approximates the behavior of a model by extracting relevant information to reveal the attribution of the input features. There are two common approaches for post-hoc explainers: \textit{model agnostic} and \textit{model specific}. The main difference is that the model agnostic explainers do not consider the model's internal components, e.g., the learned weights. Typically, they make a black box model more transparent by creating an approximation around the prediction using a linear (but local) classifier \cite{ribeiro2016lime}, feature attribution \cite{lundberg2017unified}, saliency mappings \cite{bach2015pixel,lapuschkin2019unmasking}, or rule-based explanations \cite{bastani2017interpretability}. Model agnostic approaches provide flexibility in model architecture and explanation type (e.g., linear formula or feature importance). However, they may only provide local explanations and require high computation when permuting the input features. This is problematic when dealing with large KGs that can have a huge number of entities and attributes. On the other hand, model specific explainers consider the model's internal components and mostly focus on deep neural network models. The most commonly used components are gradients \cite{liu2017visualizing,selvaraju2017grad}, attentions \cite{kumar2017explaining}, and neuron contributions \cite{chen2018learning,shrikumar2017learning,sundararajan2017axiomatic}. The limitation of this approach is that it binds to one type of model.

The methods mentioned above are designed for non-GNN models. Adapting these methods for explaining a GNN model is non-trivial, as recent studies show that these methods are prone to gradient saturation problems, especially when computing the gradient of a large adjacency matrix \cite{ying2019gnnexplainer}.

There is limited work on the post-hoc (model agnostic) explanation method of GNN models. GNNExplainer \cite{ying2019gnnexplainer} extracts the most dominant sub-graph of the input graph that affects the predictions. Specifically, the sub-graph is extracted by taking $N$ nodes that give the highest mutual information with the prediction. PGExplainer \cite{luo2020pgexplainer} improves on GNNExplainer by selectively choosing the sub-graph for the candidate explanation instead of trying all permutations of the sub-graph. GNNExplainer can highlight the top-n important neighbors in explaining the existing alignment models. However, it cannot extract the top-n important attributes since most of the existing GNN-based KG alignment models use pre-trained (or randomly initialized) embeddings as the node features.

Our proposed model fills this gap by providing a model that can accurately predict entity alignment between KGs and seamlessly provide an explanation for the prediction. The provided explanation is in the form of the similarity between the top-n features (i.e., attributes and neighbors) of the aligned entities used to compute the entity embeddings. Our model can highlight the top-n attributes and neighbors used by the model to compute the embedding of the aligned entities from two different KGs.
\color{black}

\section{Proposed KG Alignment Model}
\label{sec:proposed_model}
This section first discusses the formulation of the KG alignment problem. The detail of the proposed solution is described in the following subsection.

\begin{figure*}[t]
	\begin{center}
		\includegraphics[width=\textwidth]{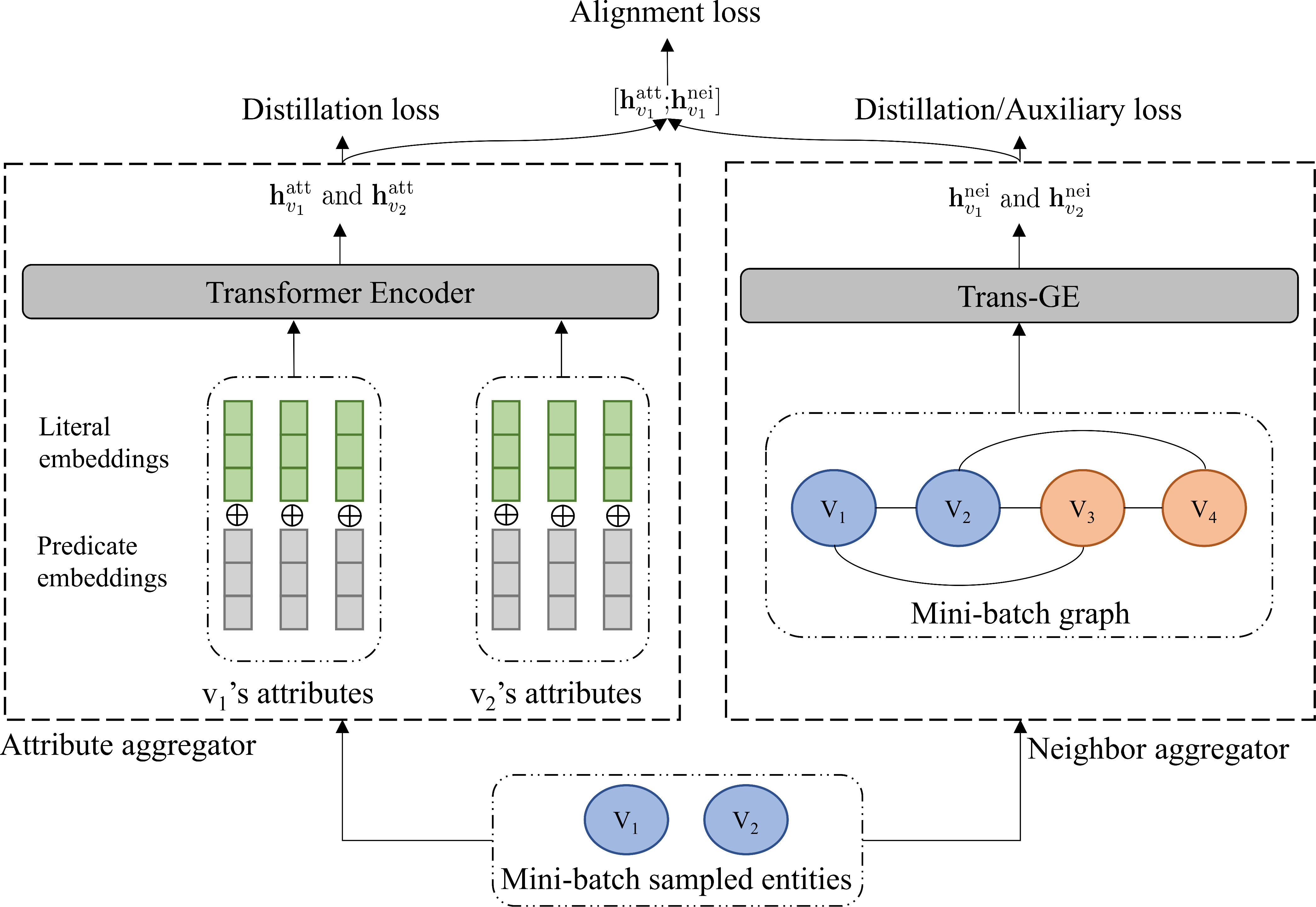}
	\end{center}
	\caption{\label{fig:overall_model} The proposed model consists of two main modules. The first is an \textbf{attribute aggregator} that aims to compute node representation based on the node's attributes. The second is a \textbf{neighbor aggregator} that aims to compute node representation based on the node's neighborhood graph structure}
\end{figure*}

\subsection{Problem Formulation}
\label{sec:problem_formulation}
A KG is an extensive repository of facts, where each fact is represented as a triple. There are two types of a triple: an attribute triple, which represents the properties of entities (e.g., \texttt{birth\_date}), and a relationship triple, which represents the relation between entities (e.g., \texttt{spouse}). An attribute triple is denoted as $\langle h,r,a \rangle$, while a relationship triple is denoted as $\langle h,r,t \rangle$. Here, $h$ is the head entity, $r$ is the predicate (i.e., the relation or the attribute key), $t$ is the tail entity, and $a$ is the attribute value. The relationship triples can be represented as a neighborhood graph $\mathcal{G}$ with vertices/nodes $\mathcal{V}$ representing entities and edges $\mathcal{E}$ representing relationships.

Let $\mathcal{G}_A$ and $\mathcal{G}_B$ denote the two KGs to be aligned. The first objective is to find every pair $\langle v_a,v_b \rangle$ where $v_a \in \mathcal{G}_A$, $v_b \in \mathcal{G}_B$ , and $v_a$ and $v_b$ represent the same real-world entity. The second objective is to provide an explanation for each extracted pair (predicted alignment). The explanation is in the form of a set of attributes and neighbors of each entity in the pair that can help experts decide whether the alignment is correct. The attributes and neighbors are selected based on their contribution in computing the entity embeddings. Here, only the most influential (e.g., top-n) attributes and neighbors are selected for the explanation. Table \ref{tab:explanation_example} shows an example of the explanation.

\subsection{i-Align}
\label{sec:i_align}
The proposed i-Align is an embedding-based KG alignment model. It uses the vector similarity of the entity embeddings to compute the similarity between entities from two different KGs. The entity embedding is computed based on attribute triples and relationship triples as follows.

\begin{align}
    \mathbf{h}_v = [\mathbf{h}_v^\text{att};\mathbf{h}_v^\text{nei}]
\end{align}
where $\mathbf{h}_v$ denotes the entity embedding, $\mathbf{h}_v^\text{att}$ denotes the attribute embeddings (a vector computed by the attribute aggregator, Section \ref{sec:attribute_aggregator}), $\mathbf{h}_v^\text{nei}$ denotes the neighborhood embeddings (a vector computed by the neighbor aggregator, Section \ref{sec:neighborhood_aggregator}), and $[;]$ denotes a concatenation operator.

The model learns the alignment via learning close entity embeddings for the aligned entities. Specifically, it learns to separate the positive samples (the aligned entities in the seed alignment $\mathcal{S}$) from the negative sample (randomly generated) by a large margin. It uses a margin ranking loss $\mathcal{L}_{align}$ defined as follows.
\begin{align}
    \mathcal{L}_\text{align}&=\sum_{s\in \mathcal{S}} \sum_{s^\prime \in \mathcal{S}^\prime} \max\left(0,\left[ \gamma+  f(s)-f(s^\prime) \right] \right) \label{eq:margin_rankin_loss}\\
	f(s)&=\left\Vert \mathbf{h}_{v_a}-\mathbf{h}_{v_b} \right\Vert \\
	f(s^\prime)&=\left\Vert \mathbf{h}_{v_a}-\mathbf{h}_{v_b^\prime} \right\Vert
\end{align}
Here, $\mathcal{S}$ is a seed alignment, containing a list of aligned entity pairs from $\mathcal{G}_A$ and $\mathcal{G}_B$, i.e., $s=\langle v_a, v_b \rangle$, that acts as positive samples. $\mathcal{S}^\prime$ is the negative samples generated by randomly changing one of the entities in each aligned entity pair of the seed alignment, e.g., $v_b$ is changed into $v_b^\prime$, a random entity in $\mathcal{G}_B$. For one positive sample, i-Align takes five negative samples.\footnote{Since a KG consists of many entities, a larger negative sample may give a better alignment performance but have a trade-off in the running-time performance. Existing KG alignment models use 5-10 negative samples.} The symbol $\gamma$ denotes the margin hyper-parameter.

i-Align is built on top of the Transformer model to exploit its self-attention mechanism to highlight the most important attributes/neighbors as the explanation of the alignment prediction. The self-attention mechanism is computationally expensive and may not be feasible to be applied to a large graph. Thus, i-Align decomposes the large graph into sub-graphs (mini-batches). It uses edge-gated attention and historical embeddings (Section \ref{sec:neighborhood_aggregator}), which allows i-Align to be trained over mini-batches.

\begin{figure}[t]
	\begin{center}
		\includegraphics[width=.75\textwidth]{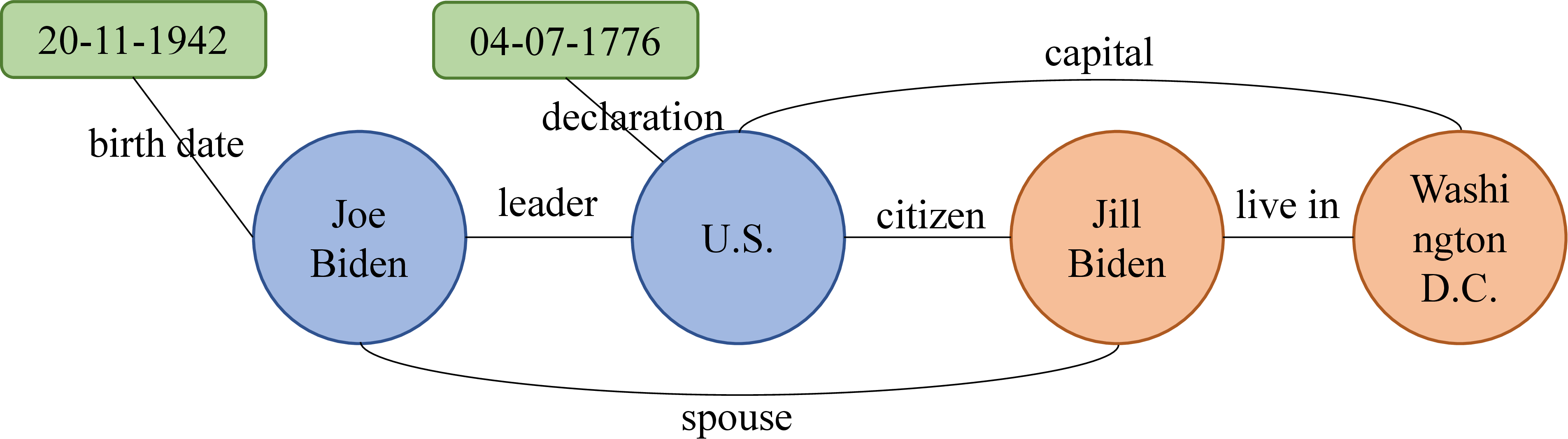}
	\end{center}
	\caption{\label{fig:mini_batch} A mini-batch sample. The blue circles indicate the initial mini-batch sample. The green rectangles indicate the attributes of the sampled entities. The orange circles indicate the first-hop neighbors of the sampled entities.}
\end{figure}

The mini-batch construction is straightforward, as illustrated in Fig. \ref{fig:mini_batch}. First, following Cluster-GCN \cite{chiang2019cluster}, a graph clustering algorithm METIS \cite{karypis1998fast} is used to sample a small number of entities that have high inter-connectivity among them (denoted by blue circles in the figure) as the initial mini-batch sample. The clustering algorithm may generate different sizes of sub-graph in the mini-batch procedure. However, this does not affect the end results of the performance. It may affect the number of epochs to reach the convergence of the model performance. If the clustering algorithm creates large graphs, it may require more epochs to reach convergence. Next, all attributes of these entities, i.e., the attribute triples, are extracted (denoted by green rectangles). These attribute triples are then sent to the attribute aggregator to be processed. Finally, all the selected entities' first-hop neighbors (denoted by orange circles), which may not be selected in the sampling process, are extracted and merged into the mini-batch. This sub-graph is then sent to the neighbor aggregator to be processed. The overall process of i-Align is illustrated in Fig. \ref{fig:overall_model}.
\color{black}

\subsubsection{Attribute Aggregator}
\label{sec:attribute_aggregator}

The goal of the attribute aggregator is to compute the attribute embeddings $h_v^\text{att}$ given the attribute triples in a mini-batch. For each entity in the mini-batch, the aggregator takes the corresponding attribute keys and values in the attribute triples. Note that the attribute value may be a literal (a sequence of character, i.e., $l = [c_1, c_2, ..., c_l]$), such as a person's name or birth date. Thus, before combining the attribute key and value to compute the final attribute embeddings, the aggregator uses Gated Recurrent Unit (GRU)\footnote{A different unit, such as LSTM, can be used here. A pre-trained encoder, such as BERT, can also be used to improve the model's capability to capture semantic similarity between attributes. GRU is chosen because it has a faster running time, with no significant drop in accuracy.\color{black}} \cite{cho2014gru} to transform an attribute value into a literal embedding (Eq. \ref{eq:char_encoder}). Next, it combines the literal embeddings and the predicate (attribute key) embeddings (Eq. \ref{eq:attribute_key_value_agg}) and uses self-attention (i.e., Transformer) to obtain the interaction between the attributes of an entity (Eq. \ref{eq:attribute_agg1}-\ref{eq:attribute_agg2}). Formally, the attribute embeddings computation is defined as follows.
\begin{align}
    \matrixize{l}_v &= \gru{\bigg([\matrixize{c}_1, \matrixize{c}_2, ..., \matrixize{c}_l]\bigg)} \label{eq:char_encoder}\\
    \matrixize{x}_v^\text{att} &= \tanh\bigg( \matrixize{W}_a \matrixize{a}_v \matrixize{W}_l \matrixize{l}_v \bigg) \label{eq:attribute_key_value_agg}\\
    \alpha_\text{att} &=\softmax\bigg(\frac{\matrixize{Q}_\text{att}\matrixize{K}_\text{att}^\transpose}{\sqrt{d}}\bigg) \quad \text{, where} \label{eq:attribute_agg1}\\
	\matrixize{Q}_a &= {\matrixize{x}_v^\text{att}}^\transpose {\matrixize{W}_\texttt{Qa}} \quad \text{and} \quad 
	\matrixize{K}_a = {\matrixize{x}_v^\text{att}}^\transpose {\matrixize{W}_\texttt{Ka}} \nonumber\\
	\matrixize{h}_v^\text{att} &= \alpha_\text{att}{{\matrixize{x}_v^\text{att}}}^\transpose {\matrixize{W}_\texttt{Va}} \label{eq:attribute_agg2}
\end{align}
Here, $\matrixize{a}_v$ is the predicate (attribute key) embeddings, $d$ denotes the dimensionality of the embedding $\matrixize{x}_v^\text{att}$ and $\matrixize{W}$ denotes learned parameters. The symbol $\alpha_\text{att}$ denotes the attention weight of the attributes, which is used to generate the explanation. The attention aggregator uses three layers of self-attention (Eq. \ref{eq:attribute_agg1}-\ref{eq:attribute_agg2}) with residual connections. Thus, the final attention weight used for explaining the attribute importance is the average of the attention weight from all layers.

\subsubsection{Neighbor Aggregator}
\label{sec:neighborhood_aggregator}

The neighbor aggregator uses a Trans\-former-based Graph Encoder (Trans-GE) to encode a mini-batch sub-graph. Two main components of Trans-GE are edge-gated attention and historical embeddings. The edge-gated attention captures the structural information of a sub-graph. At the same time, the historical embeddings allow Trans-GE to approximate the full computational graph of a KG in a mini-batch.

\textbf{Edge-gated Attention} takes the adjacency matrix and the predicate (relation) embedding of the mini-batch sub-graph to learn the attribution of entity neighbors. It multiplies the adjacency matrix and the relation embeddings and applies a sigmoid function to learn each relation's weight (Eq. \ref{eq:edge_gate}). This weight is used as a mask/gate $\gamma$ to control the neighbors' influence (Eq. \ref{eq:neighborhood_input}) on the self-attention computation (Eq. \ref{eq:neighbor_agg1}-\ref{eq:neighbor_agg2}), which also captures the similarity between relations. The weight is also used to update the relation embeddings for computing the gate in the next layer (Eq. \ref{eq:relation_update}). Specifically, the edge-gated attention is defined as follows.
\begin{align}
    \gamma &= \sigmoid\bigg({\matrixize{A} \matrixize{W}_r \matrixize{r}_v} \label{eq:edge_gate}\bigg)\\
    \matrixize{x}_v^\text{nei} &= \gamma  \matrixize{x}_\text{HE} \label{eq:neighborhood_input}\\
    \alpha_\text{nei} &=\softmax\bigg(\frac{\matrixize{Q}_\text{nei}\matrixize{K}_\text{nei}^\transpose}{\sqrt{d}}\bigg) \quad \text{, where} \label{eq:neighbor_agg1}\\
	\matrixize{Q}_n &= {\matrixize{x}_v^\text{nei}}^\transpose {\matrixize{W}_\texttt{Qn}} \quad \text{and} \quad 
	\matrixize{K}_n = {\matrixize{x}_v^\text{nei}}^\transpose {\matrixize{W}_\texttt{Kn}} \nonumber\\
	\matrixize{h}_v^\text{nei} &= \alpha_\text{nei}{{\matrixize{x}_v^\text{nei}}}^\transpose {\matrixize{W}_\texttt{Vn}} \label{eq:neighbor_agg2}\\
	\matrixize{r}_v^\prime &= \matrixize{r}_v + \gamma \matrixize{W}_{r^\prime} \matrixize{r}_v \label{eq:relation_update}
\end{align}
Here, $\matrixize{A}$ is the adjacency matrix, $\matrixize{r}_v$ and $\matrixize{r}_v^\prime$ denote the current and updated relation embeddings, respectively, $\matrixize{x}_\text{HE}$ are the initial node embeddings from the historical embeddings, and $\matrixize{W}$ denotes a learned parameter. The symbol $\alpha_\text{nei}$ denotes the attention weight of the neighbors used to generate the explanation. Similar to the attribute aggregator, the neighbor aggregator uses three layers of edge-gated attention. The final attention weight used for explaining the neighbor importance is the average attention weight from all layers.

\textbf{Historical Embeddings} are used by Trans-GE to compute the node embeddings $\matrixize{x}_\text{HE}$ (Eq. \ref{eq:neighborhood_input}), which is an approximation of node embeddings that capture the whole computational graph in a KG, i.e., it approximates the node embeddings computation of the message-passing mechanism in a GNN. It borrows the idea of GNNAutoScale \cite{Fey2021gnnautoscale} that defines the historical embeddings as node embeddings acquired in the previous training iteration, capturing the computation graph. However, the Trans-GE historical embeddings differ from those in GNNAutoScale's as they use an approximation layer instead of the push/pull mechanism used by GNNAutoScale. 

The push/pull mechanism of GNNAutoScale updates the historical embeddings whenever it computes new node embeddings in each layer so that in the next layer, it can pull the up-to-date node embeddings from historical embeddings. In contrast, the approximation layer of Trans-GE uses a linear transformation learned to approximate the current node embeddings based on the embeddings in the previous state. The advantage of using this approximation layer is that the back-propagation process of the training is more stable since it does not have disconnections in the computational graph due to the push/pull mechanism. The approximation layer is trained via distillation loss defined as follows.
\begingroup
\allowdisplaybreaks
\begin{align}
    \matrixize{x}_\text{HE} &= \matrixize{W}_\text{dist} \matrixize{x}_0 \\
    \mathcal{L}_{\text{HE}_1} &= \sum_{v \in \mathcal{G}_A \cup \mathcal{G}_B} \sum_{k=1}^{K} \bigg[1 - \cos( \matrixize{x}_{\text{HE}, v}^k,  \matrixize{h}_v^{\text{nei}, k})\bigg] \\
    \mathcal{L}_{\text{HE}_2} &= \sum_{v \in \mathcal{G}_A \cup \mathcal{G}_B} \bigg[1 - \cos( \matrixize{x}_0,  \matrixize{h}_v^\text{att})\bigg] \label{eq:distill_attribute}
\end{align}
\endgroup
where $\matrixize{x}_\text{HE}$ is the embeddings approximation, $\matrixize{x}_0$ denotes the stored historical embeddings that updated every iteration, $\mathcal{L}_\text{HE}$ is the distillation loss, and $K$ is the number of layers in Trans-GE. Following GNNAutoScale, $\mathbf{L}_2$ regularization is applied to each output layer $\matrixize{h}_v^{\text{nei}, k}$ to ensure the closeness of historical embeddings approximated in each layer. Lastly, another distillation loss (Eq. \ref{eq:distill_attribute}) is added to initialize the stored historical embeddings. The key idea is that the historical (node) embeddings can be approximated from attribute embeddings rather than random initialization.

\section{Experiments}
\label{sec:experiments}

Three experiments are conducted to show the power of i-Align.\footnote{The code and datasets are available at https://github.com/cruiseresearchgroup/i-align} The first two experiments aim to show the effectiveness of i-Align in entity alignment and explanation generation. The last experiment aims to show the scalability of i-Align to handle large KGs.

\subsection{KG Alignment Experiments}
\label{sec:exp_alignment}

Given two KGs, the main objective of i-Align is to provide entity alignment prediction and explanation. Thus, the first experiment is conducted to measure the performance of i-Align in entity alignment compared to the state-of-the-art KG alignment models. Both translation-based and GNN-based approaches are used as baselines for this experiment. Among the translation-based approaches, MTransE \cite{chen2017mtranse}, JAPE \cite{sun2017cross}, MultiKE \cite{zhang2019MultiKE} and AttrE \cite{trisedya2019entity} are selected. MTransE and JAPE are the pioneers of the translation-based approach, while MultiKE and AttrE are state-of-the-art. Among the GNN-based models, GCN-Align \cite{wang2018cross}, EPEA \cite{wang2020knowledge}, MRAEA \cite{mao2020mraea}, and NMN \cite{wu2020neighborhood} are selected. GCN-align is the pioneer of the GNN-based approach, while MRAEA and NMN are state-of-the-art. We used the suggested hyperparameters of the baseline model from their papers.

\subsubsection{Experiment Settings}
\label{sec:exp_alignment_setting}

\textbf{Dataset:} The main dataset used for the experiments is DWY-NB \cite{zhang2021comprehensive}, which is used to evaluate the alignment performance and the explanation quality. This dataset contains two pairs of KGs: DBpedia - Wikidata (\textbf{DW-NB}) and DBpedia - YAGO (\textbf{DY-NB}). The DBpedia KG in DW-NB contains $84,911$ entities and $545$ predicates, while the Wikidata KG contains $86,116$ entities and $703$ predicates. There are $50,000$ aligned entities in DW-NB. The DBpedia KG in DY-NB contains $58,858$ entities and $211$ predicates, while the YAGO KG contains $60,228$ entities and $91$ predicates. There are $15,000$ aligned entities in DY-NB. For training, $30\%$ of the aligned entities are used as seed alignment $\mathcal{S}$. The rest of the aligned entities are used for testing. We also use two additional datasets with different domains than the DW-NB dataset to further evaluate the alignment performance: \textbf{DBP-LGD} and \textbf{DBP-GEO}. These datasets contain aligned entities between DBpedia and LinkGeoData~\cite{stadler2012linkedgeodata} and Geonames\footnote{http://www.geonames.org/ontology/}, respectively. DBP-LGD contains 10,000 aligned entities and ten aligned predicates from a total of 510 predicates, and DBP-GEO contains 10,000 aligned entities and ten aligned predicates from a total of 716 predicates.\color{black}

\textbf{Hyper-parameters:} We use grid search to find the best hyper-parameters for the models. We choose the network dimensionality among $\{128, 256, 512\}$, the character embedding dimension among $\{16, 32, 64, 128\}$, the attribute and entity embedding dimension among $\{128, 256, 512\}$, the number of transformer layer among $\{2, 3, 4, 5\}$, the number of multi-head attention among $\{4, 8, 12\}$, and the margin $\gamma$ among $\{1, 5, 10\}$.

The optimum hyper-parameters are as follows. Network hidden dimension is $256$, character embedding dimension is $64$, attribute and entity embeddings dimension are $256$, the number of layers of the attributes (Transformer) and neighbor aggregators (Trans-GE) is $3$, the number of heads in the multi-head attention is $8$, margin loss $\gamma = 1$. When using a lower number than the one listed here, the performance of the model significantly drops. Meanwhile, using a higher number does not increase the performance significantly. Thus, we chose these hyper-parameters to get optimal results while having a good running time performance. The model performance stabilizes after $400$ epochs of training.
\color{black}

\textbf{Metric:} The metric used to evaluate the performance is Hits@k, e.g., \textbf{Hits@1} and \textbf{Hits@10}. It is the standard evaluation metric for KG alignment \cite{zhang2021comprehensive}. Hits@k indicates the percentage of entities with the correct aligned entity listed in the top-k prediction.

\subsubsection{Results}
\label{sec:exp_alignment_results}

\begin{table}[t!]
  \centering
  \caption{Comparisons of KG alignment models performance.}
      \resizebox{0.95\textwidth}{!}{%
        \begin{tabular}{l|cc|cc|cc|cc}
            \toprule
            \multirow{2}[4]{*}{Model} & \multicolumn{2}{c|}{DW-NB} & \multicolumn{2}{c|}{DY-NB} & \multicolumn{2}{c|}{LGD-DBP} & \multicolumn{2}{c}{GEO-DBP} \\
        \cmidrule{2-9}          & Hits@1 & Hits@10 & Hits@1 & Hits@10 & Hits@1 & Hits@10 & Hits@1 & Hits@10 \\
            \midrule
            MTransE & 7.88  & 25.75 & 0.08  & 0.68  & 33.59 & 35.76 & 33.14 & 34.75 \\
            JAPE  & 12.57 & 19.96 & 1.4   & 3.27  & 33.47 & 34.42 & 33.35 & 34.27 \\
            GCN-Align & 24.76 & 48.52 & 24.36 & 53.43 & 48.57 & 52.74 & 46.12 & 51.32 \\
            MRAEA & 81.54 & 85.97 & 73.71 & 78.52 & 78.98 & 83.13 & 72.11 & 75.32 \\
            NMN   & 84.03 & 88.21 & 75.87 & 80.54 & 78.88 & 82.35 & 75.87 & 80.18 \\
            MultiKE & 84.06 & 90.05 & 84.97 & 90.84 & 83.12 & 90.55 & 79.33 & 85.22 \\
            AtrrE & 87.98 & \textbf{95.8} & 90.44 & \textbf{94.23} & 84.17 & 92.05 & 86.91 & 92.32 \\
            Proposed & \textbf{88.35} & 94.22 & \textbf{91.21} & 93.44 & \textbf{87.21} & \textbf{94.22} & \textbf{88.87} & \textbf{93.87} \\
            \bottomrule
        \end{tabular}%
      }
  \label{tab:main_results}%
\end{table}%

Table \ref{tab:main_results} shows the comparisons of i-Align with the baseline. MTransE, JAPE, and GCN-Align achieve lower performance (i.e., below $25\%$ in terms of Hits@1 ) since they overlook entities' attributes. Specifically, MTransE and GCN-Align ignore the attributes, while JAPE masks the attributes into their corresponding data types (e.g., integer, string, etc.). MRAEA and NMN exploit the attribute label (among the other entity attributes) and achieve substantial improvements. This shows that the attribute label is an important feature for the task. \eat{\fs{To actually show that the attribute label is the most significant (or at all statistically significant) feature, wouldn't that require an ablation study, since other things also change between the various models?} \ds{Or, if statistical significance tests were not performed, replace ``significant'' with ``substantial''.}} A more detailed discussion is provided in Section \ref{sec:exp_explanation}.

The top three models are MultiKE, AttrE, and i-Align. Specifically, i-Align achieves the highest score among them: $89.42$ and $92.14$ in terms of Hits@1 on DW-NB and DY-NB datasets, respectively. The three models exploit all entities' attributes, which boosts their performance. AttrE has a slightly higher Hits@10 score as it uses a more complex n-gram model as opposed to GRU used by i-Align for computing the literal embeddings. The proposed i-Align has a further advantage compared to MultiKE and AttrE. It can also generate an explanation for each alignment prediction to help experts decide its correctness. This can help maintain the high quality of the resultant KG in the KG enrichment process.

\subsection{Alignment Explanation Experiments}
\label{sec:exp_explanation}

This experiment aims to evaluate i-Align in generating alignment prediction explanations. Here, the explanation is in the form of a set of the most influential attributes and neighbors for computing the entity embeddings. Specifically, top-5 attributes and neighbors that have the highest attention weight computed by the attribute aggregator (Section \ref{sec:attribute_aggregator}) and neighbor aggregator (Section \ref{sec:neighborhood_aggregator}), respectively, are extracted as the explanation.

\subsubsection{Experiment Settings}
\label{sec:exp_explanation_setting}
\textbf{Baseline:} GNNExplainer \cite{ying2019gnnexplainer}, the state-of-the-art explanation model of GNN, is used as a strong baseline. The explanation generated by GNNExplainer is in the form of a small sub-graph with the highest mutual information. Here, GNNExplainer is coupled with GCN-Align \cite{wang2018cross}. GCN-Align is chosen because it purely uses GCN \cite{kipf2017semi} so that it can be straightforwardly combined with GNNExplainer. This combination is adapted to compute five neighbors with the highest mutual information to compute the entity embeddings for alignment prediction. We emphasise that the combination of GCN-Align and GNNExplainer can only generate the neighbor set, while i-Align can generate both the attribute and neighbor sets.

\textbf{Protocol:} For this experiment, the dataset used is DW-NB. Fifty random samples for correct and incorrect prediction are collected (100 samples in total). We managed to get three annotators. The three annotators have studied artificial intelligence and machine learning for over ten years. They also closely work with knowledge graphs in the last three years. These three annotators are given two binary (yes/no) questions for each sample. 

For the first question, the annotators are given an attributes/neighbors set (the explanation generated by the models, Table \ref{tab:explanation_example}), and they need to indicate whether this set belongs to a correctly aligned entity pair. The second question asks the annotators whether they are sure with their answer to the first question. Three models compared for this experiment are GCN-Align+GNNExplainer, i-Align with only neighbors set explanation, and the proposed full i-Align. Each annotator is given the same 300 samples (randomly ordered), i.e., 100 samples for each model. The data given to the annotators is only the explanation, with no indication which model the sample belongs to. The metrics used for this evaluation are precision and confidence. Precision is the percentage of the correct answers (i.e., whether the given set belongs to a correct aligned entity pair) made by the annotators, while confidence is the number of answers where the annotators are sure of their answers. Fleiss' Kappa score is used to show the inter-rater agreement.
\color{black}

\begin{table}[t!]
  \centering
  \caption{Manual Evaluation of Alignment Explanation}
    \resizebox{0.9\textwidth}{!}{%
        \begin{tabu}{l|cc|cc}
        \toprule
        \multicolumn{1}{c|}{\multirow{2}[4]{*}{Model}} & \multicolumn{2}{c|}{Correct Prediction} & \multicolumn{2}{c}{Incorrect Prediction} \\
        \cmidrule{2-5}          & Prec.  & Conf.  & Prec.  & Conf. \\
        \midrule
        GCN-Align \cite{wang2018cross}& \multirow{2}[1]{*}{0.39} & \multirow{2}[1]{*}{0.47} & \multirow{2}[1]{*}{0.80} & \multirow{2}[1]{*}{0.75} \\
        + GNNExplainer \cite{ying2019gnnexplainer} &       &       &       &  \\
        \tabucline[0.5pt black!20 off 1pt]{-}
        i-Align  & \multirow{2}[0]{*}{0.68} & \multirow{2}[0]{*}{0.71} & \multirow{2}[0]{*}{0.87} & \multirow{2}[0]{*}{0.81} \\
        (Neighbors only) &       &       &       &  \\
        \tabucline[0.5pt black!20 off 1pt]{-}
        i-Align & \textbf{0.95} & \textbf{0.90} & \textbf{0.93} & \textbf{0.93} \\
        \bottomrule
        \end{tabu}%
    }
  \label{tab:alignment_explanation_manual}%
\end{table}%

\subsubsection{Results}
\label{sec:exp_explanation_results}

% \begin{table}
\begin{sidewaystable}
  \centering
  \caption{Example of Attributes/Neighbors Explanation of A \textbf{Wrong} Alignment Prediction. Both methods give a wrong alignment result of the entity \textit{Carl Ferdinand Cori}. GCN Align + GNNExplainer provides the top-5 neighbors considered by the model to make the alignment. i-Align provides both top-5 attributes and neighbors considered by the model to make the alignment. The top-5 attributes provided by i-Align make the manual check easier.}
    \resizebox{\textwidth}{!}{%
        \begin{tabular}{c|l|l|l|l}
        \toprule
              & \multicolumn{2}{c|}{i-Align Prediction} & \multicolumn{2}{c}{GCN-Align + GNNExplainer  Prediction} \\
        \midrule
        \multicolumn{1}{l|}{Entities} & Carl Ferdinand Cori & Ferdinand I of Bulgaria & Carl Ferdinand Cori & Tomáš Cihlár \\
        \midrule
        \multirow{5}[1]{*}{Attributes} & (given\_name, Ferdinand) & (given\_name, Ferdinand) & \multicolumn{1}{c|}{\multirow{5}[1]{*}{N/A}} & \multicolumn{1}{c}{\multirow{5}[1]{*}{N/A}} \\
              & (date\_of\_death, 1984-10-20) & (date\_of\_death, 1948-09-10) &       &  \\
              & (married, 1920) & (position\_end\_time, 1918) &       &  \\
              & (last\_name, Cori) & (last\_name, Bulgaria) &       &  \\
              & (date\_of\_birth, 1896-12-05) & (date\_of\_birth, 1861-02-26) &       &  \\
        \midrule
        \multirow{5}[1]{*}{Neighbors} & (occupation, biochemist) & (occupation, entomologist) & (occupation, biochemist) & (occupation, biochemist) \\
              & (citizenship, Czechoslovakia) & (native\_language, German) & (language, English) & (employer, Gilead Sciences) \\
              & (place\_of\_death, Cambridge) & (place\_of\_birth, Vienna) & (gender, male) & (country\_of\_citizenship, Czech Republic) \\
              & (member\_of, Royal Society) & (place\_of\_burial, St. Augustine's Church) & (citizenship, Czechoslovakia) & (language, English) \\
              & (place\_of\_birth, Prague) & (member\_of, Academy of Sciences) & (employer, Harvard University) & (gender, male) \\
        \bottomrule
        \end{tabular}%
    }
  \label{tab:explanation_example}%
\end{sidewaystable}
% \end{table}%

Table \ref{tab:alignment_explanation_manual} shows the results. Overall, i-Align achieves $0.95$ and $0.90$ in terms of precision and confidence, respectively, which are the highest scores among the tested models. The inter-rater agreement score is $0.81$ based on Fleiss' Kappa, which indicated high agreement between the annotators. Providing both the influential attributes and neighbors benefits i-Align, which is shown by the substantial improvement compared to when only the neighbors set of i-Align or even the baseline (i.e., the combination of GCN-Align and GNNExplainer) is used. Table \ref{tab:explanation_example} shows the top-5 attributes/neighbors of a wrong alignment prediction (i.e., i-Align aligns the entity \texttt{Carl Ferdinand Cori} with \texttt{Ferdinand I of Bulgaria}, while GCN-Align aligns it with \texttt{Tomáš Cihlár}). Here, i-Align provides a more comprehensive explanation by providing both the attribute and neighbor alignment. In contrast, GCN-Align+GNNExplainer can not provide attribute alignment as they only use structural information for computing the entity alignment. The comprehensive explanation by i-Align makes the curation process (i.e., deciding whether the given sample is an incorrect alignment prediction made by the model) a lot easier.

\subsubsection{Discussion}
\label{sec:exp_explanation_discussion}

To further analyze the alignment explanation, a semi-automatic evaluation is performed as follows. For each tested model, 100 random correct alignment predictions are collected along with their explanations (a set of influential attributes/neighbors). Next, the explanations are pre-processed by manually changing the attributes/relations and attribute values/entity tails that refer to the same meaning, e.g., the attribute \texttt{full\_name} and \texttt{name} are changed into \texttt{name}. Finally, Jaccard similarity is used to compute the set similarity. A higher score means more of the same attributes/neighbors listed in the set.

\begin{table}[t!]
  \centering
  \caption{Automatic Evaluation of Alignment Explanation}
    \resizebox{0.5\textwidth}{!}{%
        \begin{tabu}{l|c|c}
        \toprule
        \multicolumn{1}{c|}{\multirow{2}[4]{*}{Model}} & \multicolumn{2}{c}{Jaccard Similarity} \\
        \cmidrule{2-3}          & \multicolumn{1}{c}{Attributes} & Neighbors \\
        \midrule
        GCN-Align \cite{wang2018cross} & \multirow{2}[1]{*}{N/A} & \multirow{2}[1]{*}{0.26} \\
        + GNNExplainer \cite{ying2019gnnexplainer} &       &  \\
        \tabucline[0.5pt black!20 off 1pt]{-}
        i-Align & \textbf{0.48} & \textbf{0.34} \\
        \bottomrule
        \end{tabu}%
    }
  \label{tab:alignment_explanation_auto}%
\end{table}%

The results in Table \ref{tab:alignment_explanation_auto} confirm the manual evaluation where i-Align provides a better explanation than the baseline. The results show that i-Align achieves a higher set similarity score, meaning that i-Align shows more aligned attributes/neighbors to explain a correct alignment prediction. This can help experts to decide the correctness of the prediction easily. Further, i-Align is more efficient in runtime performance. In the KG alignment setting, the combination of GCN-Align and GNNExplainer is expensive. It needs to compute all the permutations of neighbors to find the maximum mutual information. Moreover, for each permutation, it needs to recompute the embeddings and the alignments.

\subsubsection{i-Align Behavior}
\label{sec:exp_behavior}

Another analysis is performed to investigate the behavior of i-Align, specifically, the effect of removing attributes/neighbors. The proposed model is run five times, and for each run, the most influential attributes (or neighbors) are selected based on their attention weight. For the next run, the selected attributes (or neighbors) are removed from the entities, and the entity embeddings are computed based on the remaining attributes (or neighbors) for predicting the alignments. All attributes (or neighbors) are removed for the last run, i.e., the model will run based on attributes or neighborhood only. For analysis, 100 aligned entity pairs that are correctly predicted in all runs are collected. This is done to see the change of feature importance used to make the correct prediction. The metrics used for this evaluation are  Hits@1 (computed based on all test data) and Jaccard similarity score (computed based on the sampled test data).

\begin{figure}[t!]
	\begin{center}
		\includegraphics[width=.75\textwidth]{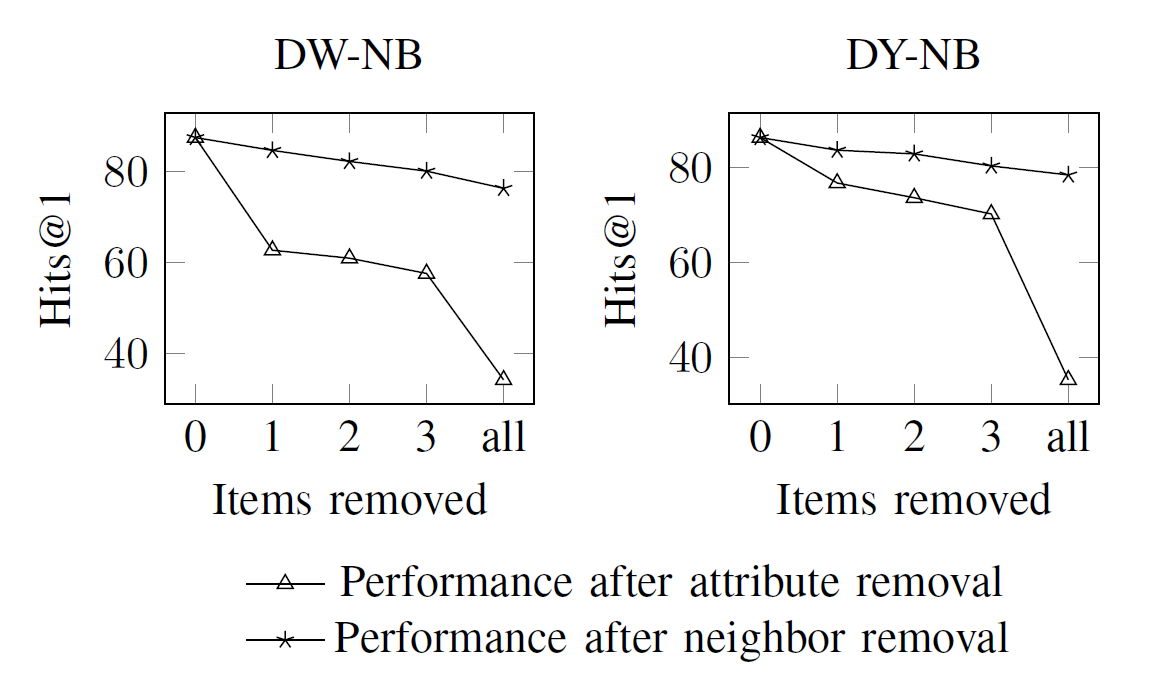}
	\end{center}
	\caption{The Effect of Attributes/Neighbors Removal on Alignment Performance}
	\label{fig:hit@1_feature_removal}
\end{figure}

\begin{figure}[t!]
	\begin{center}
		\includegraphics[width=.75\textwidth]{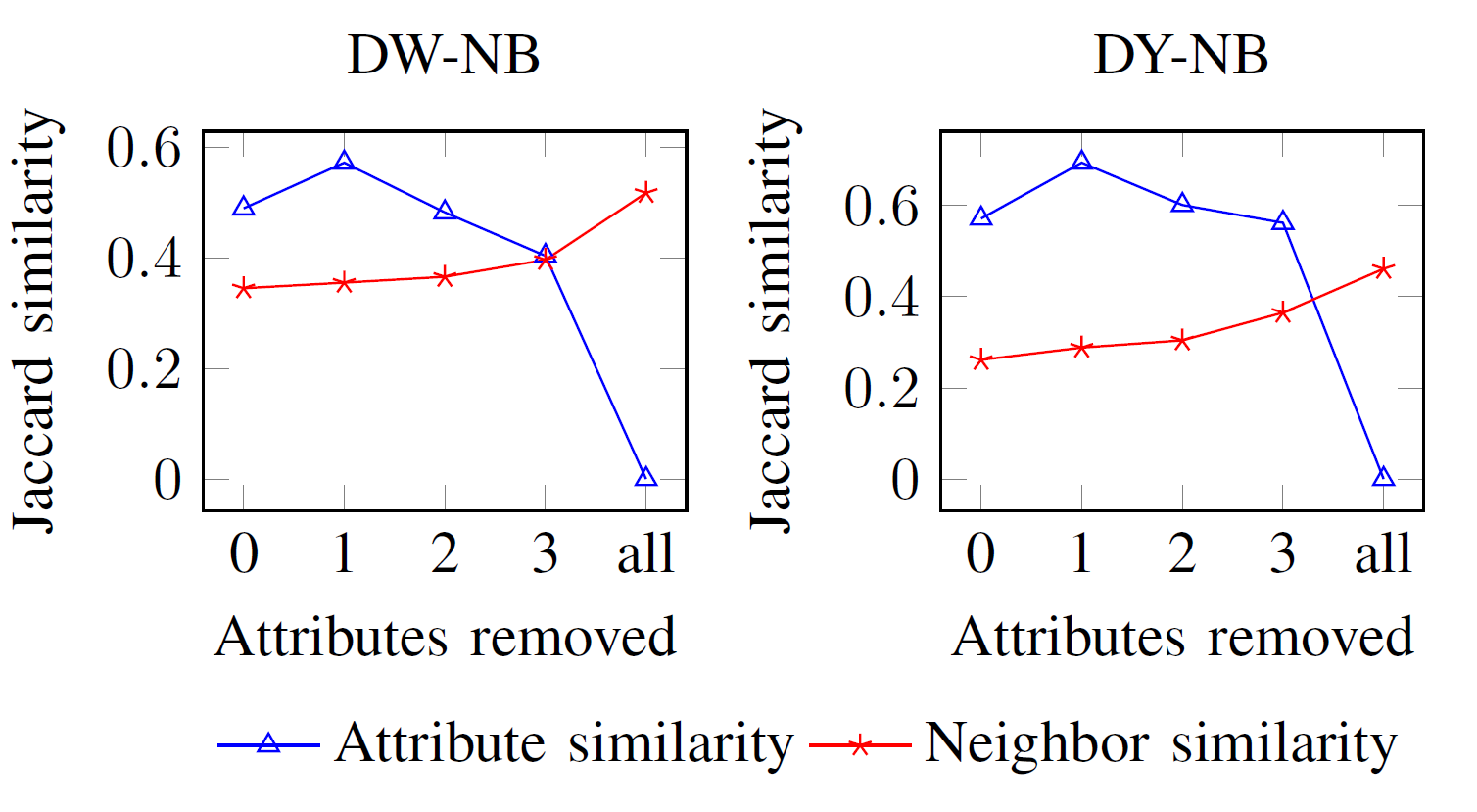}
	\end{center}
	\caption{The Effect of Attribute Removal on Set Similarity}
	\label{fig:attribute_removal}
\end{figure}

\begin{figure}[t!]
	\begin{center}
		\includegraphics[width=.75\textwidth]{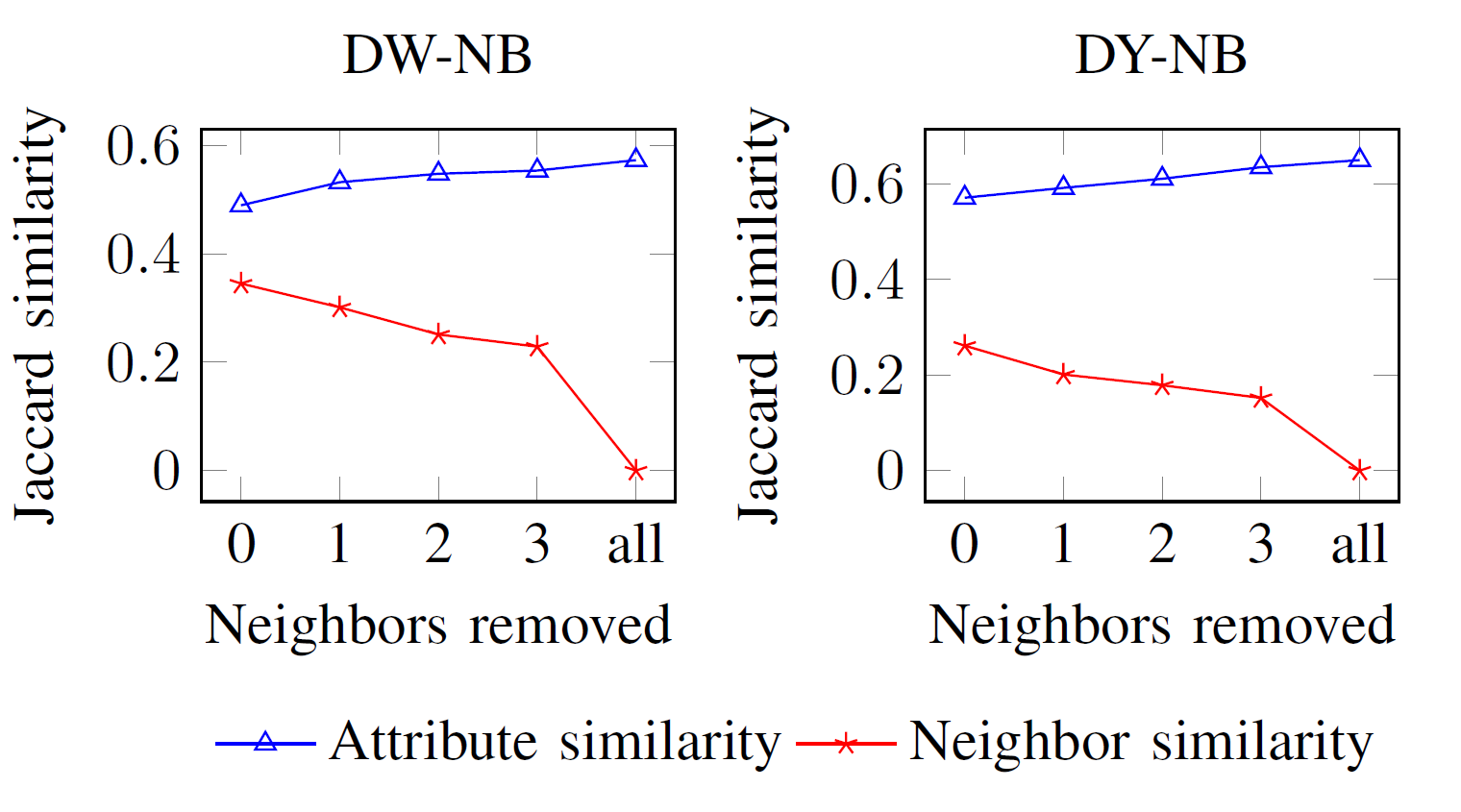}
	\end{center}
	\caption{The Effect of Neighbor Removal on Set Similarity}
	\label{fig:neighbor_removal}
\end{figure}

Figure \ref{fig:hit@1_feature_removal} shows the effect of attributes/neighbors removal on the model's performance for entity alignment. As expected, removing influential attributes/neighbors decreases the model's performance. The removal of attributes is more significantly affects the model's performance. Removing all attributes drops the performance below $40\%$. In comparison, the model's performance is still above $70\%$ when removing all the neighbors. The most significant drop occurs in the second run when removing the most influential attributes. In most cases ($97\%$), the most influential attribute is the entity name. This is in line with recent findings \cite{zhang2021comprehensive}, which show that entity name becomes a dominant feature in KG alignment.

Figures \ref{fig:attribute_removal} and \ref{fig:neighbor_removal} show the effect of attributes/neighbor removal on the importance of attributes/neighbors  by the model. In general, removing an influential attribute increases the model's attention to neighborhood similarity and vice versa. Interestingly, removing entity name, which is the dominant feature, increases both the attribute and neighbor similarity, i.e., more aligned attributes/neighbors found in the generated explanation. This means the model tries to find more evidence when there is no dominant feature.

\subsection{Scalability}
\label{sec:exp_scalability}

The last experiment evaluates the scalability of the proposed model. GNN-based KG alignment models have been shown to be effective in capturing both attributes and neighbors/graph structural similarity. However, the existing GNN-based models are constrained by memory usage when aligning large KGs. The proposed i-Align handles this problem by using historical embeddings.

The datasets used for this experiment are DW300K and DW600K \cite{zhang2021comprehensive}, which are three times and six times bigger than DW-NB, respectively. The baselines for this experiment are AttrE \cite{trisedya2019entity} and MultiKE \cite{zhang2019MultiKE}. They belong to the translation-based models. No GNN-based models are used as a baseline since they cannot run on the machine used for experiments because of out-of-memory error. The workstation used for this experiment has a 2.20GHz processor, 128GB main memory, and a GPU with 32GB memory.

\begin{table}[t!]
  \centering
  \caption{Scalability Experiments Results}
    \resizebox{0.75\textwidth}{!}{%
        \begin{tabular}{l|cc|cc}
        \toprule
        \multicolumn{1}{c|}{\multirow{3}[4]{*}{Model}} & \multicolumn{2}{c|}{DW300K} & \multicolumn{2}{c}{DW600K} \\
        \cmidrule{2-5}          & \multirow{2}[2]{*}{Hits@1} & Running Time & \multirow{2}[2]{*}{Hits@1} & Running Time \\
              &       & (days) &       & (days) \\
        \midrule
        AttrE \cite{trisedya2019entity} & 70.59 & 5.6   & 61.22 & 10.9 \\
        MultiKE \cite{zhang2019MultiKE} & 69.56 & 6.0   & 61.42 & 11.5 \\
        i-Align & \textbf{72.26} & 5.9   & \textbf{63.37} & 11.2 \\
        \bottomrule
        \end{tabular}%
    }
  \label{tab:experiment_scalability}%
\end{table}%

Table \ref{tab:experiment_scalability} shows the scalability experiments result. The proposed i-Align achieves better performance than the baseline with a reasonable running time. The results show the power of i-Align in the effectiveness and practicality of aligning large KGs. Specifically, this confirms the effectiveness of the historical embeddings of i-Align. Note that the baseline is Translation-based models that cannot produce explanations for  alignment prediction.

\section{Conclusion and Future Work}
\label{sec:conclusion}

This paper proposed i-Align, an interpretable KG alignment model. The main advantage of i-Align over the existing KG alignment models is that it provides an explanation for each alignment prediction made. This explanation can help experts in the curation process to merge KGs by providing an explanation of proposed alignment predictions. Thus, it helps maintain the high quality of the enriched KG. The proposed model has two components: attribute aggregator and neighbor aggregator. The attribute aggregator uses the standard Transformer, while a novel Transformer-based Graph Encoder (Trans-GE) is proposed for the neighbor aggregator. Trans-GE uses \textit{Edge-gated Attention} that combines the adjacency matrix and the self-attention matrix to learn a score as a gate to control the information aggregation from the neighboring entities. It also uses \textit{historical embeddings}, allowing Trans-GE trained over mini-batches/small sub-graphs to address the scalability issue when encoding a large KG. The attention mechanisms of the attribute and neighbor aggregators are used to compute the attention weight to highlight the important attributes and neighbors, respectively. Experimental results show the model's effectiveness for aligning KGs, the quality of the generated explanations, and the practicality for aligning large KGs.

The proposed i-Align uses attention weights as the primary indicator of the importance of attributes/neighbors. This is a simple yet effective technique. One limitation of i-Align is that it uses attribute literal similarity. So, it may not perform well on cross-lingual knowledge graph alignment, where the literal attributes are written in different characters. The other interesting topics for future work are: an integration to a more advanced technique, such as attention rollout \cite{chefer2021rollout}, which can be explored to improve the explanation quality; explainability in GNNs is an emerging research topic and worth exploring to improve i-Align.
\color{black}

\section{Acknowledgement} 
\label{acknowledgement}
This research is supported by Australian Research Council (ARC) Centre of Excellence for Automated Decision-Making and Society (\textit{ARC CE200100005}).

\bibliography{reference.bib}
\end{document}